# An Optical Frontend for a Convolutional Neural Network

Shane Colburn[1, +], Yi Chu[2, +], Eli Shlizerman[1,2, *], Arka Majumdar[1,3, *]

[1]Electrical and Computer Engineering, University of Washington, Seattle, WA-98195
[2]Applied Mathematics, University of Washington, Seattle, WA-98195
[3]Department of Physics, University of Washington, Seattle, WA-98195
[+]Two authors contributed equally
[*]Email: shlizee@uw.edu, arka@uw.edu

**Abstract:** The parallelism of optics and the miniaturization of optical components using nanophotonic structures, such as metasurfaces present a compelling alternative to electronic implementations of convolutional neural networks. The lack of a low-power optical nonlinearity, however, requires slow and energy-inefficient conversions between the electronic and optical domains. Here, we design an architecture which utilizes a single electrical to optical conversion by designing a free-space optical frontend unit that implements the linear operations of the first layer with the subsequent layers realized electronically. Speed and power analysis of the architecture indicates that the hybrid photonic-electronic architecture outperforms sole electronic architecture for large image sizes and kernels. Benchmarking of the photonic-electronic architecture on a modified version of AlexNet achieves a classification accuracy of 87.1% on images from the Kaggle's Cats and Dogs challenge database.

**Introduction:** Artificial Neural Networks (ANNs) with a deep layered structure have shown advanced capabilities for solving ubiquitous large-scale computational problems in recent years [1-4]. In particular, Convolutional Neural Network (CNN) architectures have enabled superior performance over alternative approaches in classification and pattern recognition problems in computer vision [5-9]. In these applications, the input image is convolved with kernels of various dimensions, and the outputs of the convolutions are subsequently pooled, passed through a nonlinear activation function, and then directed to successive convolutional layers [10]. While CNNs boost performance in terms of the ability to solve classification and recognition problems, they require a large number of computations, primarily due to the vast computational requirements of convolution operations with large images and kernels. The total number of computations rapidly becomes prohibitive with an increasing number of layers (the depth of the network) and input size (number of pixels). Convolving an input image containing $n \times n$ number of pixels with a kernel of shape $k \times k$ yields a computational complexity of $O(n^2 k^2)$ [11]. This creates a significant bottleneck that results in a high latency and large power consumption even for unidirectional propagation (forward inference) in a pre-trained network. Software implementations of these networks realized sequentially are impractical to use for large images and datasets. While latency can be reduced substantially with dedicated electronic hardware and the significant parallelism offered by graphics processing units (GPUs), the computation time and energy consumption still preclude real-time inference, see Figure 2 [12].

Free-space optical elements are known to be very efficient linear processors of spatial information [13] in terms of energy and speed. For example, a lens can passively perform a two-dimensional Fourier transform in the brief time (picosecond scale) it takes light to travel twice the focal length of a lens. In contrast, when this task is implemented electronically it has the complexity of $O(n^2 log(n))$, $n^2$ being the total number of pixels, which makes it a slow and power hungry procedure. In free-space optics, the well-known 4f correlator architecture exploits the Fourier transform property of a lens to perform arbitrary convolutions very efficiently. This had previously motivated researchers to explore implementing neural networks using light [14, 15].

Unfortunately, many free-space optical implementations of these networks relied on macroscopic and bulky refractive elements, leading to large sizes and strong alignment sensitivities, hindering widespread adoption of the platform [16].

With the advent of nano-patterned subwavelength diffractive optics, commonly known as metasurfaces, it is now possible to realize optical elements in a flat and compact form factor [17-20], and the problems associated with size and misalignment in previous optical implementations of ANNs can be circumvented. Metasurfaces comprise spatially-varying arrays of subwavelength-spaced optical antennas that can impart transformations in amplitude, phase, and polarization on incident electromagnetic waves [21, 22]. These devices have already enabled a class of flat optical elements, including visible wavelength implementations of lenses [23, 24], vortex beam generators [25], holograms [26], blazed gratings [27], and freeform optical surfaces [28, 29]. Coupled with the ability to fabricate metasurfaces using well-established semiconductor fabrication tools, there is great potential for realizing the next generation of miniaturized elements for free-space optical information processing.

Even with the promise of miniaturization and the drastic reduction in alignment sensitivity via monolithic nanofabrication of layered metasurfaces, there are still outstanding challenges to realizing an efficient and scalable optical neural network. Two of the most significant of these challenges are the inability to tune metasurfaces arbitrarily pixel-by-pixel and the lack of a low-power nonlinearity in the optical domain. While several groups have demonstrated all-optical switching at extremely low power [30-32], realizing multiple switches in a network has not been demonstrated. Similarly, ultra-low power electro-optical modulators exist [33, 34], but designing an array of such modulators for processing a large amount of data in parallel remains elusive. While the thermo-optic effect has been used to train integrated photonics-based ANNs before [35], the large energy consumption in these heaters limits the efficiency of the network.

A promising approach for realizing an optical neural network (ONN) is to augment the optical hardware with electronic implementations of the nonlinearity and to use a pre-trained network [36, 37], bypassing the need to dynamically tune metasurfaces as training would be conducted offline. Combining optics and electronics, however, presents its own challenges. Specifically, the requirement of converting a large amount of data between the optical and electronic domains is costly both in terms of energy and latency. For deep networks with many layers, the necessity of repeatedly converting back and forth between these signal domains would limit any speedup provided by optics as signal transduction is slow. The energy cost would also be prohibitive. For convolutional neural networks, however, where often the preponderance of the computational burden is allocated to the initial layers, the use of optics could be justified. In the limit of only the first layer of a CNN being implemented optically, only a single signal conversion step is required. We call this first layer an optical frontend. In this paper, we design and simulate such a network, where an optical frontend coupled with electronic implementations of successive layers is used to implement a CNN. Our design leverages metasurface optics to implement 4f-correlator-based Fourier filtering operations and we benchmark its performance by realizing an optical frontend for AlexNet [6], a widely adopted and state-of-the-art CNN for object classification. We analyze our network's classification accuracy and speed as a function of input image size, benchmarking against a fully software-based version of AlexNet [38].

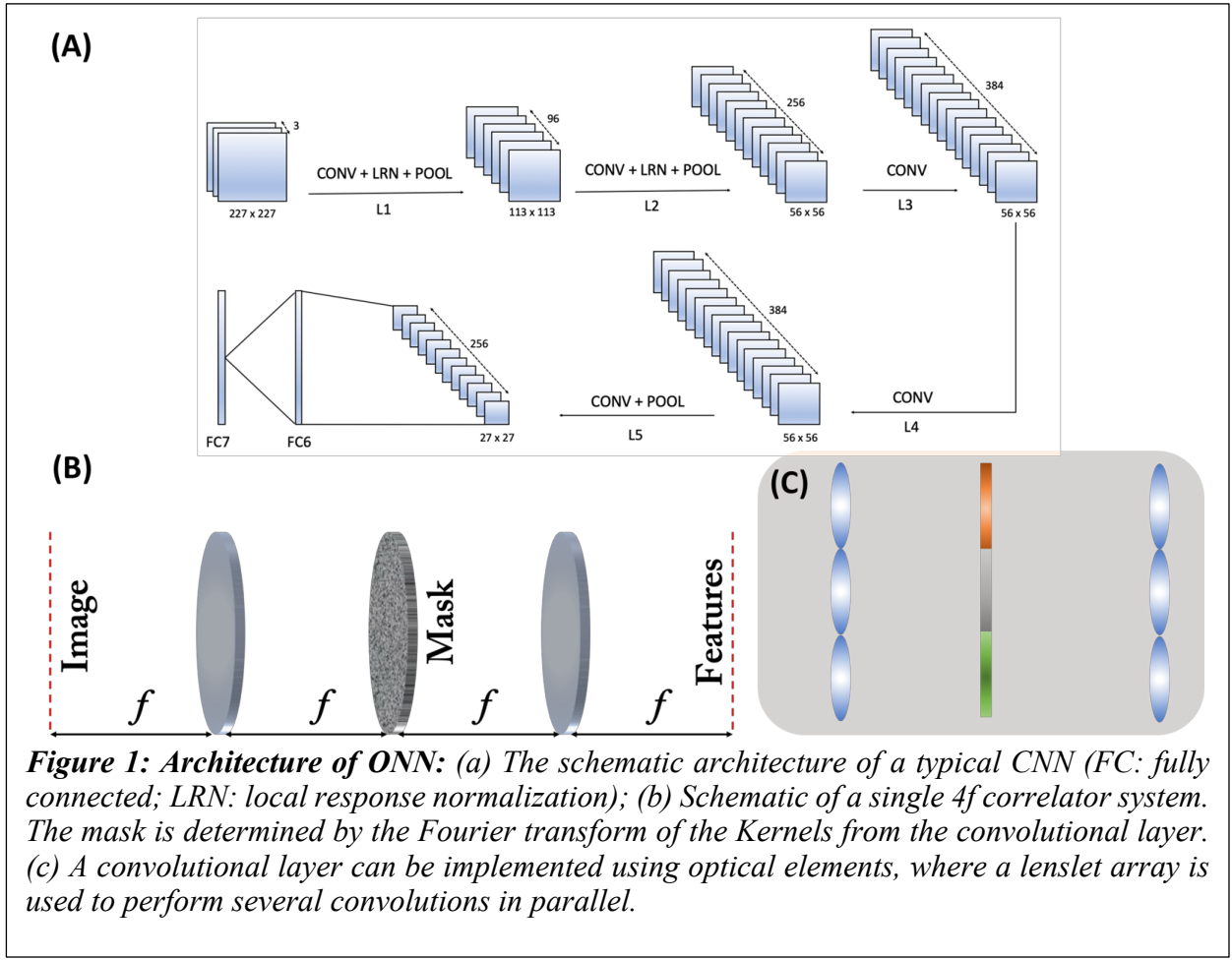

*Figure 1: Architecture of ONN: (a) The schematic architecture of a typical CNN (FC: fully connected; LRN: local response normalization); (b) Schematic of a single 4f correlator system. The mask is determined by the Fourier transform of the Kernels from the convolutional layer. (c) A convolutional layer can be implemented using optical elements, where a lenslet array is used to perform several convolutions in parallel.*

**Architecture of the optical CNN:**
A CNN consists of a sequence of layers in which sets of kernels are convolved with an input image, the outputs of the convolutions are pooled together, and nonlinear thresholds are applied to subsequent images (Figure 1a). By performing multiple convolution operations in parallel on an image in a layer, multiple outputs are generated per layer. These are then directed to pooling operators that coalesce these outputs into a smaller set of images for propagation to the next layer where they undergo convolutions with a new set of kernels. Via successive action of the convolution, pooling, and nonlinear operations, higher dimensional features of images can be extracted and identified to enable high accuracy discrimination between different types of objects present in a set of input images. In this section, we detail our architecture for realizing a CNN based on an optical frontend unit which implements the first set of convolution operations optically with the remaining layers of the network realized electronically.

Our architecture leverages the traditional 4f correlator design, comprising two lenses of equal focal length spaced apart by $2f$ and with input and output planes located at the front and back focal planes of the first and second lenses respectively (Figure 1b). Within the Fresnel approximation, as a lens provides an exact Fourier transform relation between the electric field distributions located at its front and back focal planes, two lenses in series provide a sign-flipped Fourier transform of the electric field at the input. With a mask positioned in the plane between the two lenses, arbitrary Fourier domain filtering operations can be implemented, including low, high, bandstop, bandreject, and more exotic filters if complex-valued transmittance masks are used in the filter plane [39].

As CNN requires many convolution operations, our architecture comprises an array of 4f correlators to enable all the convolution operations of a single layer to occur in parallel. To implement this structure, we propose a system comprising a stack of two aligned lenslet arrays with an array of filter masks inserted between (Figure 1c) them. To ensure compactness, such a system could be realized using flat optics based on visible regime efficient metasurfaces [18, 23]. Stepper lithography-based fabrication could enable high-throughput and scalable fabrication of such structures [29, 40] and the masks could be separated by either polymer spacer layers (e.g., using SU8) or free-space by integrating them in a precision-printed 3D housing [41]. To transmit and post-process images in our system, we require an array of sources aligned with our lenslet array as well as an array of sensors to collect the output from each 4f correlator in the lenslet stack. The choice of visible wavelength operation is motivated by the availability of cheap laser-sources and arrays of silicon photo-detectors. We design for 3 different coherent sources, one for each color channel of red (632 nm), green (532 nm), and blue (442 nm).

In the Fourier transform plane between the pair of lenses for each 4f correlator, masks can be placed to perform convolution with a desired kernel. These masks are the Fourier domain equivalent of the pre-trained spatial kernels used in the convolutional layer. As the Fourier domain filters are complex-valued (containing both phase and amplitude information), they cannot be efficiently implemented via binary amplitude masks. To realize a general complex-valued two-dimensional mask, we propose to use a phase metasurface, leveraging techniques used previously in phase-only spatial light modulators (SLMs) to implement general complex-valued functions [42]. This technique utilizes an enlarged pixel size comprising a checkerboard pattern of subpixels that switch between two phase values, creating an averaging effect that is the vector sum of the two values, enabling access to any polar coordinate on or within the unit circle (i.e., any phase from 0 to $2\pi$ and any amplitude between 0 and 1). This comes at the cost of enlarged pixel sizes, but with the subwavelength spacing of metasurface scatters, the spatial resolution will still be significantly improved relative to SLM-based implementations. In the plane of the mask, a point-by-point multiplication between the Fourier transform of the object and the Fourier domain filter occur inherently as the light transmits through the complex-valued transmittance mask. The resulting distribution then undergoes an inverse Fourier transform by the second lens, projecting the spatial domain output of the convolution onto its back focal plane. The convolution results are then converted to electrical signals using a nonlinear square law photodetector. This electrical signal can then be passed to subsequent layers in the network that are implemented in software or can be converted back to the optical domain for another set of convolution operations.

The lenses used in our 4f architecture are 0.57 mm wide with a focal length of 3 mm. At these dimensions, the lenses have sufficiently low numerical aperture (NA) such that they can be regarded as paraxial, making the Fourier transforming property of a lens valid. The paraxiality of the system will be an important consideration when attempting to scale this system to a much lower volume [43]. If the NA of the lenses in the array becomes too high, the Fourier transforming property will not hold as the Fresnel approximation will break down. Furthermore, with smaller apertures, the fundamental information capacity of the system (i.e., the space-bandwidth product) will decrease, which limits the number of channels, or pixels, for information processing. The space-bandwidth product of an imaging system is given by $\left(\frac{D^2}{\lambda f}\right)^2$, with $D$ being the aperture dimension, $f$ the focal length and $\lambda$ the optical wavelength [44]. For our system with wavelengths between $\lambda \sim 400 - 650 nm$, the space-bandwidth product becomes $\sim 200 \times 200$, implying the system can reliably perform computation over $\sim 200 \times 200$ pixels.

To simulate the optical portion of our hybrid system, we use a custom wave optics-based code. We use the fast Fourier transform (FFT) algorithm and the angular spectrum propagator to model diffraction and calculate the electric field distribution at different propagation distances. This model only assumes that the input wave can be treated as a scalar and makes no assumptions about the paraxiality of the system, making it more general than techniques based on the Fresnel propagator. For each convolution kernel, we perform a split-step simulation, alternating between real and Fourier domain spaces as we propagate through free-space and impinge on optical elements (i.e., lenses or filters). We model our lenses and filters as complex amplitude masks that we can multiply elementwise with the incident electric field. To model the photodetection, we take the magnitude squared of the electric field and apply a proportionality constant which will depend on the exposure time and responsivity of the particular detector used in experiment.

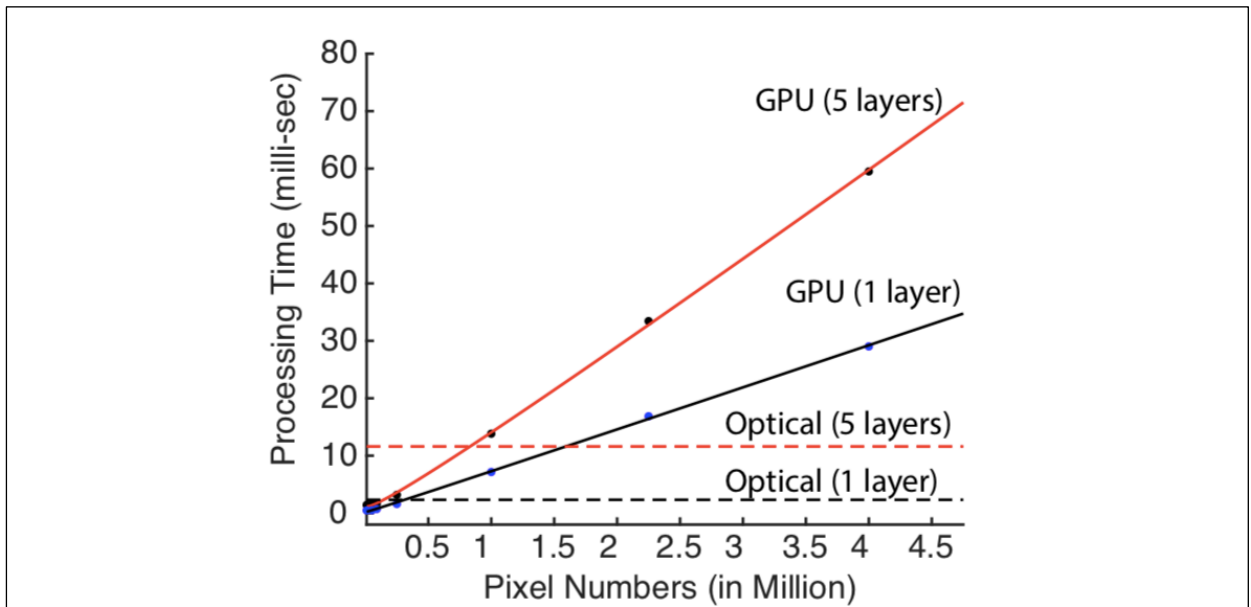

*Figure 2: Comparison of total processing time between AlexNet GPU (red solid-all 5 layers; black solid-1st layer only) and OPCNN (red-dashed-all 5 layers; black dashed-first layer only) as a function of pixels number ($n^2$). Processing time (y-axis milliseconds) for AlexNet GPU increases linearly with image size (x-axis $n^2$). OPCNN processing time is estimated to stay constant for various image sizes, with the bulk of the time coming from the signal transduction time. We find that processing time becomes favorable for OPCNN with image-size about 250 thousand pixels (500 x 500) to 1 million pixels (1000 x 1000).*

Due to the limited GPU hardware and memory available to us, it was not feasible to simulate the whole diffractive lenslet array as a single optical element. Hence, for computational ease, to model the system, we simulate each 4f correlator separately and sequentially. This approximation assumes that there is negligible crosstalk between each 4f correlator. If a significant portion of light from one correlator enters adjacent correlators, then such a simulation method would be inaccurate. Hence, coupling between each lens stack in the array would need to be assessed, and to evaluate the validity of our method, we calculated the crosstalk between lenses in our system. Using the final design parameters for our focal length (3 mm), lens diameter (0.57 mm), convolution kernels, and pixel pitch (2.5 μm), we simulate a single 4f correlator that is centered in a 3 × 3 array of spaces, where each tile in the array has the same aperture as of the 4f correlator's lenses (i.e., each 4f correlator is placed directly adjacent to its neighbors, with no spacing between). We inject a representative object pattern (an image of a poodle) into the centered 4f correlator and then calculate the fraction of incident power that is then distributed into the output plane of the surrounding 8 tiles in the 3 × 3 array. This fraction captures the essence of how much light in one channel leaks into adjacent correlators in the array. By repeating this simulation for all the kernels used in our convolutional layer and for all color channels, we found that our average crosstalk was 0.0084, or less than 1% of light leaks from the center 4f correlator into the surrounding 8 correlators. This justifies neglecting the crosstalk in modelling our system, enabling us to treat each 4f correlator as an independent convolutional unit.

The negligible crosstalk between our 4f correlators implies that the convolutional units can be packed into dense arrays without any spacing between them, enabling significant size reduction compared to the case where crosstalk is nonnegligible. To implement the distinct 96 kernels for each of the 3 color channels (red, green, and blue) of the first convolutional layer of AlexNet, for 227 × 227 resolution images this would require a ~0.94 $cm^2$ array of 4f correlators. As the area required for the optical frontend is proportional to the number of kernels, for networks with fewer kernels than AlexNet, the area can be reduced significantly.

**Speed, Energy and Complexity Analysis:** A significant portion of the energy consumption and processing time in our ONN comes from the signal transduction step. The latency in our system for a single optical convolutional layer ($T_{latency}$) is the sum of the time to generate a new input image ($T_{source}$), for the light to propagate through the lenslet stack ($T_{4f}$), be detected by the CCD array ($T_{detect}$), and then be transmitted ($T_{data}$) for subsequent software processing as indicated in the equation below

$$T_{latency} = T_{source} + T_{4f} + T_{detect} + T_{data}.$$

Our sources will consist of SLMs that can be refreshed at 1 kHz frequency, resulting in images being generated as fast as 1 ms. The detection time depends on the responsivity of the detector and the input power level but based on the available technology, the latency can be estimated to be 1 ms. If the CCD's image data is then transmitted via USB 3.0 protocol at a rate of 2500 Mbit/sec and assuming a 100 kB image, the data transmission step requires 0.32 ms. As we design our optical elements using compact metasurfaces, which can achieve very short focal lengths (we design our lenslet arrays such that each lens is 0.57 mm in diameter with focal lengths of 3 mm), the light propagation time is very short, taking only *~10 ps*. Thus, the *total latency* associated with a single convolutional layer is *2.32 ms*.

The overwhelming majority of the time required to perform the convolutions comes from the source generation, CCD sensing, and data transmission, with the actual convolution step itself in the 4f correlator ($T_{4f}$) having a negligible time contribution. Figure 2 compares the forward computation time for several different hardware implementations of the first few layers of AlexNet, including the times for computing the first layer optically and electronically, and for computing all 5 convolutional layers optically and electronically. To obtain an accurate benchmark for the electronic times, we averaged the computation time of 100 forward runs. While the electronic layers' computation time increases linearly with number of pixels, the optical implementation's times remains constant for any image size as the duration depends only on the fixed time associated with source generation, photodetection, and data transmission, since the optical convolution time itself is negligible. This also indicates that unless the number of image pixels is large, optical convolution does not provide a significant time benefit. This is where use of a free-space implementation of the ONN becomes important due to the large required space-bandwidth product, as an integrated photonic realization of the convolutional layer would require the same number of waveguides as the number of pixels, posing a serious limitation.

To evaluate the effectiveness of using optical convolution for a CNN, we also estimate how the components in the AlexNet architecture contribute to forward computation time. The architecture of AlexNet is composed of 5 convolutional layers, where the first two layers (*L1, L2*) include the sequential operations of *convolution, ReLU, local response normalization (LRN)*, and *max-pooling*. The third and the fourth layers *(L3, L4)* consist of *convolution* and *ReLU* only, whereas the fifth layer *(L5)* includes *convolution, ReLU,* and *max-pooling [6]*. The convolution operation utilizes a set of filters (tensors) and during forward inference, each input is convolved with the filters, the elements of which are learnable parameters updated during training. The ReLU operation applies a nonlinear elementwise activation function. It is followed by LRN in the first two layers. The max-pooling operation reduces the number of parameters by selecting the maximum elements from slices of the output of the nonlinear operation. Convolution is the most computationally complex and time-consuming of these operations, whereas normalization and max-pooling are an order of magnitude faster. The exact computation time for convolution depends on the dimension and number of kernels, as well as the dimension and number of input images.

***Table 1:*** *Computational cost of different layers in AlexNet: Computation times (in milliseconds) of each layer in AlexNet estimated on a CPU. The first two layers are more time consuming taking up more than 60% of the processing time (L1:25.1%, L2:37.6%). In this simulation, the number of pixels used is $227 \times 227$.*

| Layer# | 1 | 2 | 3 | 4 | 5 | Total |
|---|---|---|---|---|---|---|
| **Inference time (ms)** | 2.75 | 4.11 | 1.39 | 1.55 | 1.15 | 10.9 |
| **% of the total time** | 25.1% | 37.6% | 12.6% | 14.2% | 10.5% | 100% |

We estimated the sequential forward computation time and deconstructed it into the portions of time it takes to complete the computation of each layer. We summarize our results in Table 1. Layers 1 and 2 are more time consuming than the subsequent layers, together constituting 62.7% of the total time with Layer 2 being the most computationally expensive. This is consistent with the number of operations performed in each layer. We estimate the computation time on a CPU (Intel i7 8 core), as when the calculation is performed in parallel on a GPU (NVIDIA-TitanX), we cannot separate and analyze the computation time layer by layer. GPU forward time is up to 7× faster than on a CPU (GPU time is 1.52 ms), see Figure 2 and [45]. From this analysis and given the time bottleneck of optical-electrical signal transduction, we validate our restriction of replacing a single convolutional layer with an optical implementation to limit the number of fixed transduction time delays. The first layer is the optimal selection because while it is only the second most time consuming, it is also the layer that processes the initial input image that could potentially already exist in the optical domain, thus requiring only a single optical to electrical conversion at the detector side.

We also analyzed the total energy required by the neural network. The convolution operation in the optical domain does not consume any excess energy, and thus the energy consumed in the optical implementation primarily depends on the signal transduction. To estimate the energy, we start from the detector side, where the optical signal strength will depend upon the noise floor. Assuming ~$1\mu W$ power at the detector side per pixel, that each optical element transmits a fraction $t$ of the incident power, and that the source efficiency is $\eta$, the total optical power required is

$$P_{optical} = \frac{n^2 \times n_{kernel}}{\eta \times t^p} \mu W$$

with $n^2$ being the total number of pixels per 4f correlator, $p$ being the number of optical elements in the path, and $n_{kernel}$ being the number of different kernels used in the convolutional layer. For an electronic implementation, each operation will require a certain amount of energy ($P_{switching}$), and the total energy will be

$$P_{electronic} = \alpha \times n^2 \times k^2 \times n_{kernel} \times P_{switching}$$

where the constant α is a coefficient determined by the architecture on which the implementation is executed. The energy scaling of the optical and electronic implementations shows that both scale in the same manner with the number of pixels and number of kernels for convolution. For the optical implementation, however, the power is independent of the size of the kernel, whereas for electronics it is not. As such, for large kernels, an optical convolutional layer offers reduced power consumption.

**Classification Accuracy:** To compute the accuracy of our hybrid ONN, we benchmarked its performance along with the other convolutional layers (*L2-L5*) of AlexNet implemented using standard computation (software running on a GPU). For the optical frontend layer, we implemented a simulation of the optical system in the TensorFlow-Python framework and connected it with a TensorFlow-Python implementation of the remaining layers [46]. We chose the Kaggle's Cats and Dogs classification challenge as our benchmark [47] and divided the data in the challenge (37.5K images) into 3 sets: training (30K images), validation (2.5K images), and test (5K images) sets. We first estimated the classification accuracy of the AlexNet network using pre-trained weights on the ImageNet database [6, 38, 46]. We loaded the pre-trained weights and trained only the fully connected layer of AlexNet on our training set. The classification accuracy of this network on the test set was 96.4% (Table 2-C1). When we replaced the first layer with our optical frontend, however, using the pretrained weights from AlexNet to realize our complex-valued Fourier domain transmittance masks, the accuracy dropped to 49.98% (Table 2-C2). This accuracy (near 50%) indicates that the classification task does not perform well and is closer to the performance of a random classifier.

This reduction in accuracy is expected since there are several major differences between AlexNet and our own network, such that the pretrained weights are no longer valid because the network structure and operations are fundamentally different. One major difference is that the convolutions performed by the optical frontend are the effective convolutions performed by the 4f correlators, which is equivalent to continuous domain convolution, and a close approximation to discrete domain convolution where the kernel is shifted in increments of 1 pixel for each successive multiplication and sum. In AlexNet, however, the convolutions of the first layer were pretrained using a standard choice of a stride of 4 (i.e., the kernel is shifted in intervals of 4 pixels instead of 1). Indeed, strides other than 1 are not achievable in the optical domain. Another significant difference is the optical frontend's use of a square nonlinearity, unavoidable because of the intensity response of a photodetector, instead of the ReLU nonlinearity of AlexNet. Furthermore, the optical frontend does not include a constant bias offset term (i.e., an addition operation at the end of the layer) as in AlexNet [10].

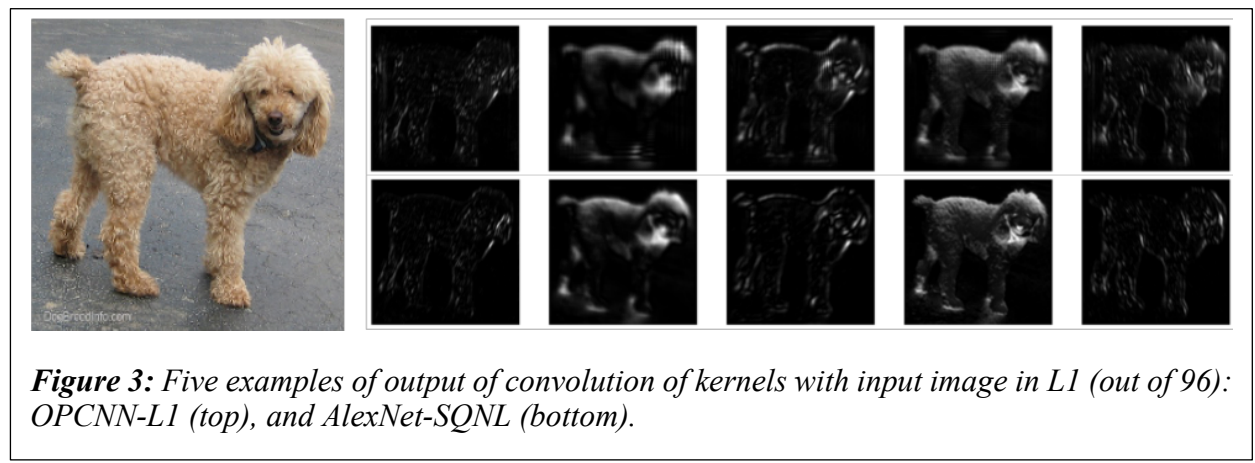

**Figure 3:** *Five examples of output of convolution of kernels with input image in L1 (out of 96): OPCNN-L1 (top), and AlexNet-SQNL (bottom).*

We modified the AlexNet to accommodate these differences (i.e., changed the network to perform convolution with a stride of 1, replaced the ReLU nonlinearity with a square function for the first layer, and removed the bias applied in the first layer) and then performed training and validation with our respective datasets. We refer to this new network architecture as AlexNet-SQNL as it uses a square nonlinearity for the first convolutional layer. We configured the training process to stop and estimate the classification accuracy when the rate of change in validation accuracy and the cross entropy are below $10^{-3}$. For AlexNet SQNL, we obtained 87.3% classification accuracy and set this as the 'ground truth' for the comparison with the optical frontend-based network (Table2-C4). We then train our optical frontend-based network (OPCNN-L1) with the same training and validation sets and stopping criterion as for AlexNet SQNL. Notably, the training time for each epoch is much longer (x3.2 times) than for AlexNet SQNL since the TensorFlow-Python simulation of the optical frontend in software entails a split-step simulation of light propagation via the angular spectrum method, equivalent to solving the Rayleigh-Sommerfeld diffraction integral [39]. When implemented in hardware, the computation time for the optical layer would be negligible compared to the other components of the system, the simulation time only reflects the computational complexity of modeling the system in software. With the trained OPCNN-L1 network, we obtain a classification accuracy of 87.1% (Table2-C3), achieving nearly the same accuracy as AlexNet SQNL (our 'ground truth' network). In Figure 3 we show visual examples of the output of the convolution of kernels with a sample input image in *L1* for both OPCNN-L1 and AlexNet SQNL and observe that the outputs are highly similar with only minimal visual distinctions between them.

*Table 2: Classification Accuracy for different networks*

| Network | AlexNet Pretrained (C1) | OPCNN-L1 AlexNet Pretrained (C2) | OPCNN-L1 AlexNet Trained (C3) | AlexNet-SQNL (Ground Truth) (C4) |
|---|---|---|---|---|
| **Accuracy** | 96.4% | 49.98% | 87.1% | 87.3% |

**Discussion:**

While the achieved classification accuracy with our OPCNN-L1 network (87.1%) is lower than that of the pre-trained AlexNet on the ImageNet database (96.4%), the OPCNN-L1's accuracy is nearly the same as the 'ground truth', or classification accuracy of the fully software-based version (AlexNet SQNL). Because of the similarity in accuracy between AlexNet SQNL and OPCNN-L1, this is indicative not of any limitation of the optical frontend itself but rather a high bias in our network architecture introduced by changing from a ReLU to square type nonlinearity, utilizing a stride of 1 in our first layer, and not having a bias term. It is likely that the classification accuracy can be improved by altering or introducing operations into the electronic portion of the network and leveraging the speed of our optical frontend. For example, after photodetection, a bias term could be introduced and a ReLU nonlinearity or other operation could be applied to the subsequent data before proceeding to the first electronic layer, potentially enabling both improved speed via the optical frontend and comparable accuracy to the pre-trained AlexNet architecture.

Our investigation also shows that CNNs adapted for use with an optical frontend are sensitive such that variations in configuration, even seemingly minor ones, necessitate retraining. Specifically, substitution from stride 4 to stride 1 and modification of the first layer's nonlinearity from ReLU to square nonlinearity required re-training. We also found that our accuracy improves with additional data. Here, we attempted training with 15k, 20k, and 30k images and observed improvement with each increment of the data set. Of course, increasing the training set size means that more inputs will be propagated through the network, requiring a prolonged training time. With sufficient computational resources, however, training could be conducted offline in this manner to improve the accuracy of the optical frontend-based network while still enabling a speedup in inference time.

Our analysis indicates that the full benefit of using optics for neural networks cannot be achieved without a low-power optical nonlinearity. Substituting electronic nonlinearities in place of fast yet physically infeasible optical nonlinearities introduces latency bottlenecks to the data pipeline in a neural network as conversion between electronic and optical domains is very costly in terms of time and power. This bottleneck led us to consider implementing a single optical layer within a network of multiple convolutional layers to reduce the number of conversions, especially since for many CNNs the preponderance of the computational complexity is within the initial layers as they entail more convolution operations and larger image sizes and kernels. In our architecture, we specifically chose to optically implement the first layer of a CNN (an optical frontend), and our complexity analysis and benchmarking confirmed this is one of the most time-consuming layers. Furthermore, the first layer as the initial layer receives the data in the optical domain, which, eliminates the need for an extra signal conversion. While this computing paradigm does not perform as well as electronic counterparts for small images, it is well suited to applications where high-resolution images are processed. The processing time for our 4f correlators is independent of image size, since the only latency from the optical frontend arises from the electronic image generation, sensing, and data transmission, whereas for electronics the computation time increases linearly with the number of pixels.

In this paper, we described the design, simulation, and analysis of a CNN architecture based on an optical frontend unit coupled with an electronic backend. Our optical frontend comprises an array of 4f correlators with filter masks inserted in-between, with the correlators' lenses implemented as metasurfaces and simulated via planewave spectrum calculations. The frontend implements all the linear operations of the first layer, the set of convolution operations performed on the image passed to the input layer, and transmits the convolved outputs in parallel onto an array of CCDs that enable the data to be propagated through the rest of the network, which is implemented electronically. Using our proposed architecture, we evaluated its capabilities by implementing a modified version of AlexNet with it and compared its performance to the original and fully electronic-based version of the network. We achieved a classification accuracy of 87.1%, nearly the same as the 'ground truth' network that we evaluated for comparison. We anticipate that this accuracy can be improved with more training data as well as modification of the network architecture by introducing a bias term and additional nonlinearity in software after the outputs of the frontend are captured via the simulated CCDs. Our network demonstrated superior scaling capabilities compared to electronic counterparts, with no dependence on input image size for computation time or kernel size for power consumption, whereas electronic-based convolution has a computation time that scales linearly with number of pixels and power consumption that scales quadratically with kernel size. Our proposed architecture may find applications in tasks that require high-resolution images and may usher in the next generation of hybrid optical-electronic information processing units.

**Acknowledgement:** The research is funded under Samsung-GRO grant, royalty research fund (RRF) from University of Washington, Seattle, National Science Foundation grant No. DMS-1361145 (ES), and Washington Research Foundation Innovation Fund (ES). We also acknowledge useful discussion with Dr. Albert Ryou.

*Our code can be found at:* https://github.com/shlizee/OpticalNN